\documentclass{elsarticle}
\usepackage{amsmath,amsfonts}
\usepackage{algorithmic}

\usepackage{algorithm}
\usepackage{array}
\usepackage[caption=false,font=normalsize,labelfont=sf,textfont=sf]{subfig}
\usepackage{textcomp}
\usepackage{stfloats}
\usepackage{url}
\usepackage{verbatim}
\usepackage{graphicx}
\usepackage{cite}
\usepackage[utf8]{inputenc} 
\usepackage[T1]{fontenc}    
\usepackage{booktabs}       
\usepackage{multirow}
\usepackage{tabularx}
\usepackage{nicefrac}       
\usepackage{microtype}      
\usepackage{xcolor}         
\usepackage{todonotes}
\usepackage{amsthm,amssymb}
\usepackage{comment}
\usepackage{adjustbox}

\usepackage{colortbl} 

\newcommand{\R}{{\mathbb R}}

\renewcommand{\Re}{\mathbb{R}}
\newtheorem{proposition}{Proposition}
\newtheorem{lemma}[proposition]{Lemma}

\newtheorem{assumption}{Assumption}

\usepackage{algorithm,algorithmic}
\journal{Expert Systems with Applications}

\begin{document}

\begin{frontmatter}


\title{Beyond adaptive gradient: Fast-Controlled Minibatch Algorithm for large-scale optimization}

\author{Corrado Coppola\corref{cor1}}
\ead{corrado.coppola@uniroma1.it}
\cortext[cor1]{corresponding author}
\affiliation{organization={Sapienza University of Rome - Departement of Computer, Control and Management Engineering},
            country={Italy}}
\author{Lorenzo Papa}
\ead{lorenzo.papa@uniroma1.it}
\author{Irene Amerini}
\ead{irene.amerini@uniroma1.it}
\author{Laura Palagi}
\ead{laura.palagi@uniroma1.it}

\begin{abstract}
Adaptive gradient methods have been increasingly adopted by deep learning community due to their fast convergence and reduced sensitivity to hyper-parameters.
However, these methods come with limitations, such as increased memory requirements for elements like moving averages and a poorly understood convergence theory. 
To overcome these challenges, we introduce F-CMA, a Fast-Controlled Mini-batch Algorithm with a random reshuffling method featuring a sufficient decrease condition and a line-search procedure to ensure loss reduction per epoch, along with its deterministic proof of global convergence to a stationary point.
To evaluate the F-CMA, we integrate it into conventional training protocols for classification tasks involving both convolutional neural networks and vision transformer models, allowing for a direct comparison with popular optimizers.
Computational tests show significant improvements, including a decrease in the overall training time by up to $68\%$, an increase in per-epoch efficiency by up to $20\%$, and in model accuracy by up to $5\%$. 
\end{abstract}


\begin{keyword}
deep learning \sep optimization algorithms \sep computer vision
\end{keyword}

\end{frontmatter}



\section{Introduction}
\label{sec:intro}
In recent years, with the introduction of popular optimizers such as Adam \citep{Kingma2015AdamAM}, adaptive gradient methods have gained popularity in the deep learning community because of their fast convergence and lower sensitivity to hyper-parameters. 
Nonetheless, they suffer from drawbacks, including the need for a memory storage which increases linearly with the number of trainable parameters \citep{shazeer2018adafactor} and a less investigated convergence theory \citep{zaheer2019study}.
In this work, we propose F-CMA, a fast-controlled mini-batch algorithm enhanced with a derivative-free extrapolation line-search, to solve the problem of the unconstrained minimization of the sum of $P$ $L$-smooth, possibly non-convex functions $f_i:\R^n\rightarrow\R$ in the form:
\begin{equation} \label{eq:fw}
 f(w) = \sum_{i=1}^P f_i(w).
\end{equation}

In the context of machine learning (ML), $f_i(w)$ represents the loss function; precisely, in supervised ML, it represents a measure of the dissimilarity between the ground truth $y_i$ and the predicted output $\hat y_i$ for a given input sample $x_i$, where $w \in \R^n$ is the vector of the model parameters.
Recently, the exponential increase in the number of model parameters, driven by the advancement in deep learning architectures, such as attention-based and diffusion models \citep{lin2022survey, khan2022transformers, devlin2018bert, radford2018improving, vaswani2017attention}, combined with the growing adoption of over-parametrization to achieve optimal interpolation \citep{allen2019convergence, buhai2020empirical, xu2018benefits, chang2021provable}, yielded the use of traditional first-order methods\footnote{First-order methods require the full-batch gradient of the objective function to solve the problem of the unconstrained minimization of \eqref{eq:fw}.} prohibitively expensive.

Hence, the most largely adopted optimizers are the mini-batch methods, which rely on parameters updating rules that exploit only a small subset of the available dataset.
While the underlying idea of mini-batch methods dates back to the early 1950s with \citep{robbins1951stochastic}, plenty of these methods have been developed in the past decade. 
These methods aim to balance the delicate trade-off between strong convergence properties (e.g., convergence guarantees and convergence rates) and computational efficiency requirements, often imposed by hardware constraints, such as limited GPU.
Moreover, they can be categorized into adaptive and non-adaptive methods based on whether the learning rate is static across all parameters or dynamically adjusted during the training process.

Non-adaptive methods, like Stochastic Gradient Descent (SGD) \citep{robbins1951stochastic, bottou2018optimization, ruder2016overview, Sutskever2013OnTI} and its variants such as Incremental Gradient (IG) \citep{bertsekas2000gradient}, Incremental Aggregated Gradient (IAG)\citep{gurbuzbalaban2017convergence,vanli2018global}, and random reshuffling (RR) \citep{mishchenko2020random} perform parameters update with fixed or decreasing according to a pre-determined schedule learning rate $\alpha^t$ in the form:
\begin{equation}\label{eq:nonadaptive}
    w^{t+1} = w^t + \alpha^t d^t
\end{equation}
where $d^t$ is an estimate of $-\nabla f(w^t)$.
Despite being simpler to implement with respect to their adaptive competitors, these methods are highly affected by the initial learning rate setting.
The need for a careful tuning of the initial step size is crucial yet computationally intensive, as finding an optimal setting involves an exhaustive pre-training phase.
Furthermore, as highlighted in \citep{yang2024two}, on the one hand the optimal fixed learning rate can have strong dependencies on the inverse of the Lipschitz constant of $f(w)$, which can lead to extremely slow convergence.
On the other hand, adopting updating rules based on pre-determined schedule without a careful setting of the decrease rate can lead to exploding gradient and diverging loss function \citep{bengio1994learning,lecun2015deep}. 

Conversely, adaptive methods perform parameter updates using a fixed learning rate and a direction, that is, an aggregated measure (e.g., the exponential moving average) of the gradient estimates sampled until the current step.
Many adaptive methods have been proposed by deep learning researchers \citep{mcmahan2017survey,haji2021comparison}, each varying the gradient accumulation rule. 
In this work, seeking for a comparison between F-CMA and the most frequently used optimizers, we consider as benchmark the following adaptive methods: Adam \citep{Kingma2015AdamAM}, Adamax \citep{Kingma2015AdamAM}, AdamW \citep{loshchilov2017decoupled,loshchilov2018fixing}, Adagrad \citep{duchi2011adaptive}, Nadam \citep{Dozat2016IncorporatingNM}, and Radam \citep{liu2019variance}.
Despite requiring a larger amount of memory to store the past gradient estimates and their second moments \citep{shazeer2018adafactor}, adaptive methods are generally acknowledged to be more efficient, stable, and robust to the initial learning rate choice \citep{zhou2020towards,ma2022qualitative,yang2024two}. 
However, adaptive methods require stronger assumptions on the objective function to prove convergence, such as strong convexity of $f(w)$ \citep{kovalev2020optimal,hanzely2021accelerated,kovalev2021lower} or gradient-related search direction \citep{gratton2023convergence,gratton2024complexity}, which are often not satisfied in deep learning application \citep{richards2021learning,kawaguchi2016deep}.
Moreover, most of the available converge theory rely on stochastic results, i.e., the convergence is attained only in the expected value \citep{bottou2018optimization}.

In addition to non-adaptive and adaptive methods, there have been multiple attempts to address the problem of learning rate tuning in non-adaptive algorithms such as making SGD adaptive through the use of a line-search procedure \citep{jiang2024adaptive,mutschler2020parabolic,mutschler2021empirically,mahsereci2017probabilistic,gulcehre2017robust}.
Building on this research path and considering the advantages that the RR strategy has demonstrated in numerous applications compared to its incremental and stochastic counterparts  \citep{gurbuzbalaban2021random,ying2017performance,sharma2024improved,safran2020good,mishchenko2022proximal},  we propose a method that integrates RR  \citep{mishchenko2020random} (Algorithm  \ref{alg:innercycle1}) with a safeguard rule to ensure the sufficient decrease of an approximation, $\tilde{f}$, of $f(w)$. 
More in detail, if the safeguard rule is not satisfied, inspired by the controlled mini-batch framework proposed in  \citep{liuzzi2022convergence}, F-CMA automatically adjusts the learning rate by either decreasing it or executing a line-search procedure (Algorithm \ref{alg:EDFL2_cma1}), discussed in Section \ref{sec:convergence_dim}.
Similarly to \citep{coppola2023cma} (CMAL), the decrease condition on $\tilde f$ at each epoch is checked using the aggregated running loss over an epoch. 
However, F-CMA performs the line-search without the true loss function, and only two full evaluations of $f(w)$ are required (Step 3 and Step 12 in Algorithm \ref{alg:EDFL2_cma1}), reducing the computational overhead with respect to CMAL.
Summarizing, the main features of F-CMA are following reported:
\begin{itemize} 
\setlength\itemsep{0.05cm}
    \item We prove that $\inf_k \| \nabla f(w^k) \| \rightarrow 0$, allowing the random reshuffling of the samples at each epoch using a deterministic approach and without requiring neither the convexity of $f(w)$ nor a gradient-related search direction.
    \item We introduce a learning rate-driven early stopping rule to reduce the computational effort, lowering the consumption energy required for the training process and its carbon footprint.
    \item We design a derivative-free line-search technique that can use any approximation model of $f(w)$ and requires, in the worst case, at most two full evaluations of $f(w)$ while still preserving convergence properties and competitive performances.
    \item We lighten the individual coercivity assumption of \citep{coppola2023cma} on $f_i(w)$ and replace it with gradient clipping to ensure the boundedness of $\|\nabla f(w^k)\|$ and speed up the training process \citep{qian2021understanding,chen2020understanding}.
    \item We introduce an internal scheduler ($\eta$ in Algorithm \ref{alg:EDFL2_cma1}) to adjust the learning rate $\zeta^k$ before the line-search.
\end{itemize}
{Finally, the rest of the paper is organized as follows. 
In Section \ref{sec:relatedwork}, we provide a relevant literature review.
In Section \ref{sec:convergence_dim} we describe the mathematical aspects of F-CMA and discuss its convergence properties.
In Section \ref{sec:expsetup}, we explain the experimental setup used to compare the proposed algorithm with state-of-the-art optimizers, and in Section \ref{sec:results}, we investigate its estimation performances compared over six well-known architectures on two benchmark datasets.}

\section{Related Work}
\label{sec:relatedwork}
In this section, we provide a detailed description of the main existing optimization algorithms for deep learning, categorized between non-adaptive, adaptive, and line-search based methods.

\textbf{Non-adaptive methods.} Most of the available literature is focused on improving SGD convergence properties, assuming different hypotheses, such as in \citep{li2017convergence,gower2019sgd,liu2022almost,nguyen2018sgd}.
In the case of smooth objective $f(w)$ with decreasing learning rate, it is proved that $\mathbb{E}[\|\nabla f(w^k)\|] \rightarrow 0$ \citep{bertsekas2000gradient}.
Different complexity bounds can be found when assuming additional hypotheses, such as setting a learning rate decreasing rule that depends on the Lipschitz constant of $L$ \citep{bottou2018optimization,bolte2017error,ghadimi2013stochastic}.
Eventually, possible convergence proofs of RR and convergence rates are provided in \citep{mishchenko2020random,li2023convergence}.
Conversely, for both Incremental Gradient (IG) and Incremental Aggregated Gradient (IAG), it is possible to prove convergence not only in the expected value, i.e., to prove that $\inf_k \| \nabla f(w^k) \| \rightarrow 0$.
Concerning IG, its convergence is proved with smooth objective function in \citep{bertsekas2000gradient}. 
Furthermore, \citep{blatt2007convergent} provides convergence rates in case of fixed learning rate and  \citep{gurbuzbalaban2019convergence} in many other cases according to the learning rate decreasing rule.
The convergence theory of IAG is discussed \citep{gurbuzbalaban2017convergence}, while in \citep{mokhtari2018surpassing} an IAG variant is presented with linear convergence rate.

\textbf{Adaptive methods.} The underlying mathematical theory of adaptive methods is much less studied than SGD, IG, and IAG.
To the best of our knowledge, a full convergence theory has been developed only on Adam and Adagrad.
Concerning Adam, convergence proofs have been provided under different hypotheses, most of all involving second-order conditions \citep{li2024convergence} or achieving only local convergence \citep{bock2019proof} or only for some specific hyper-parameters settings \citep{zhang2022adam}. 
Similar results can also be found in \citep{ward2020adagrad,liu2023convergence,kairouz2021nearly,li2019convergence,xie2020linear} on Adagrad.

\textbf{Line-search based methods.} Recent advancements in stochastic gradient descent (SGD) have already employed line-search techniques to enhance convergence rates and stability, particularly in noisy environments and over-parameterized models. 
\citep{mahsereci2017probabilistic} introduces a probabilistic line-search method that combines deterministic techniques with Bayesian optimization, dynamically adjusting the step size to improve stability and reduce parameter tuning needs.
\citep{vaswani2019painless} proposes a stochastic Armijo line-search tailored for over-parameterized models, achieving deterministic convergence rates under interpolation conditions without increasing batch size.
\citep{gulcehre2017robust} explores Lipschitz and Armijo line-search techniques to select step sizes, ensuring decent properties without overshooting and improving convergence speed and robustness by adapting to the local landscape of the objective function. 
\citep{loizou2021stochastic} proposes a stochastic Polyak step-size, which adapts the learning rate based on current function values and gradients, ensuring efficient convergence in over-parameterized models without needing problem-specific constants or additional computational overhead.
Finally, \citep{liuzzi2022convergence,coppola2023cma} introduce the controlled mini-batch algorithm (CMA) and its approximated version CMAL. 

To summarize, most of the existing methods are proved to converge only in the expected value, i.e., $\mathbb{E}[\|\nabla f(w)\|]\rightarrow0$, and the non-adaptive ones are particularly sensitive to the choice of the learning rate.
F-CMA, is much less sensitive to the initial learning rate due to its line-search and internal step-size decreasing rule. 
Moreover, despite the cyclical reshuffling of samples, we still provide a deterministic global convergence proof in Section \ref{sec:convergence_dim}, which highlights its increased stability with respect to the main competitors.

\section{Fast-Controlled Mini-batch Algorithm}
\label{sec:convergence_dim}
In this section, we discuss in detail the proposed method and the convergence properties of F-CMA.

\subsection{Initial assumptions}

Our objective is to solve the unconstrained optimization problem reported in Equation \ref{eq:minfw}.
\begin{equation} \label{eq:minfw}
    \min_{w \in \R^n} f(w) = \sum_{i=1}^P f_i(w)
\end{equation}

Where $f_i(\cdot)$ represents any loss function, such as cross-entropy loss (CEL) or mean squared error (MSE), $w$ is the vector of a neural network trainable parameters, while the index $i=1,2,\dots, P$ is a sample in the data set or a batch of them.
In this context, $f_i(\cdot)$ is often named batch loss function.

F-CMA, reported in Algorithm \ref{alg:CMA1}, relies on checking after every epoch a sufficient decrease condition on an estimate of the real objective function.
This estimate is obtained by summing the batch losses accumulated during an epoch.
According to Algorithm \ref{alg:innercycle1}, given a random permutation $I^k$ at iteration $k$, this estimate be formally written as:
\begin{equation}\label{eq:f_tilde}
\tilde f^k = \sum_{p=1}^P f_{h_p} (\tilde w^{k-1}_{h_{p-1}})    
\end{equation}

where $\{\tilde w^k_{h_{p-1}}\}_{p=1,\dots, P}$ is the trajectory of points generated by a RR iteration, as reported in Algorithm \ref{alg:innercycle1}.

\begin{algorithm}[h!]
\footnotesize
\begin{algorithmic}[1]
    \STATE {\bf Input}: $w^k$, $\zeta^k$
    \STATE  Let $I_k = \{h_1,\dots,h_P\}$ be a random permutation of $\{1,\dots,P\}$
    \STATE  Set $\widetilde w_{h_0}^k = w^k$
    \FOR{$i =1,...,P$}
    \STATE {$\tilde f^k_{h_i}= f_{h_i}(\widetilde w_{h_{i-1}}^k)$}
        \STATE $\tilde d_{h_i}^k = -\nabla f_{h_i}^k(\widetilde w_{h_{i-1}}^k)$
        \STATE $\widetilde w_{h_i}^k= \widetilde w_{h_{i-1}}^k + {\zeta^k} \tilde d_{h_i}^k$
    \ENDFOR 
    \STATE {\bf Output} $\widetilde w^k = \widetilde w^k_P$, $d^k = \displaystyle\sum_{i=1}^P\tilde d_{h_i}^k${, $\tilde f^k=\displaystyle\sum_{i=1}^P\tilde f^k_{h_i}$}
\end{algorithmic}
\caption{\texttt{Random Reshuffling (RR)}}
\label{alg:innercycle1}
\end{algorithm}

Before proceeding to the detailed explanation of F-CMA, we introduce the assumptions we make to prove the global convergence to a stationary point.
Notice that these assumptions will be considered always satisfied and omitted from the theoretical statements for the sake of brevity.

\begin{assumption}[Coercivity of $f(w)$]\label{ass:compact}
The function $f$ is coercive, i.e., all the level sets 
\begin{equation}
  {\cal L}_f(w_0) = \{w\in\Re^n:\ f(w) \leq f(w_0)\}
\end{equation}
are compact for any $w_0\in\Re^n$.
\end{assumption}

\begin{assumption}[$L$-smoothness] \label{ass:lipschtizagradient}
$f_p(w) \in \mathcal{C}^1_{L_p}(\R^n)$ for all $p=1,\dots,P$, meaning that the functions $f_p$ are all continuously differentiable and the gradients $\nabla f_p$ are Lipschitz continuous for each $p=1,\dots, P$, namely there exists finite $L = \max_{p=1,\dots,P} L_p >0$ such that:
\begin{equation}
 \|\nabla f_p(u)-\nabla f_p(v)\|\le L\|u-v\| \forall \ u,w\in\R^n
\end{equation}
\end{assumption}

To simplify the notation of the more frequently recurring terms, for the rest of the paper and in the proofs reported in the appendix, we will refer to the Lipschitz constant of $f_p(w)$ as $L_{f_p}$, while we will indicate the constant of $\nabla f_p$ as $L_p$.

\begin{assumption}[$f_p$ bounded below]\label{ass:fiproperties}
The functions $f_p$ are bounded below for each $p=1,\dots, P$, that is:
\begin{equation}
    \label{eq:fpcoercive}
    \begin{aligned}
        \mathcal{L}_{f_p}(w_0) &= \{ w \in \mathbb{R}^n : f_p(w) \leq f_p(w_0) \} \\
        &\text{ is compact for any } w_0 \in \mathbb{R}^n.
    \end{aligned}
    \end{equation}
\end{assumption}

\begin{assumption}[Growth condition]\label{ass:nablabounded}
There exist two constants $C,D > 0$ and $M>0$ such that for each $p=1,\dots, P$ and for any $w \in \R^n$:

\begin{equation}
    \|\nabla f_p(w)\|\le C\| \nabla f(w) \| + D
\end{equation}
\begin{equation}
    \|\nabla f(w)\|\le M
\end{equation}

\end{assumption}

More in detail, assumptions \ref{ass:compact},\ref{ass:lipschtizagradient}, and \ref{ass:fiproperties} are often used in smooth optimization for deep learning and recognized as non-restrictive \citep{defossez2020simple,ward2020adagrad,bertsekas2000gradient,mishchenko2020random,sun2020optimization,zaheer2019study,solodov1998incremental}.
Moreover, Assumption \ref{ass:nablabounded}, despite being more restrictive, is still widely adopted in algorithms that use approximations of the true objective, such as optimization in federated learning with error \citep{richtarik2016distributed,fercoq2015accelerated,levy2021storm+,dorfman2024dynamic}, adaptive methods in non-convex optimization \citep{ward2020adagrad,defossez2020simple}, and stochastic gradient method (e.g., \citep{bertsekas2000gradient,allen2016variance,bottou2018optimization} assume bounded variance of $\|\nabla f(w)\|$).
We remark that the convergence still holds if Assumption \ref{ass:nablabounded} is satisfied for any $w$ belonging to sequence of points $\{w^k\}$ produced by F-CMA.
In the computational practice this condition can be enforced using the gradient clipping option\footnote{\url{https://pytorch.org/docs/stable/generated/torch.nn.utils.clip_grad_norm_.html}}, which is also acknowledged to speed up the training process when using non-adaptive methods \citep{chen2020understanding,qian2021understanding}.

The first proposition below is what will allow us, at the end, to proof global convergence. 
In fact, with the following result, we admit that F-CMA can converge to a stationary point if the sequence of points produced is limited and the learning rate goes to zero.

\begin{proposition}[Convergence assuming $\{w^k\}$ bounded]
\label{prop:limitRR}
Assume that the sequence of points $\{w^k\}$  produced by Algorithm \ref{alg:innercycle1} is limited and that $\lim_{k\rightarrow0}\zeta_k = 0$. 
Then, for any limit point $\Bar w$ of $\{w^k\}$ a subset of indices $K$ exists such that 
    \begin{eqnarray}
        && \lim_{k\to \infty,k\in K} w^k=\Bar w,\label{assert_C}\\
        && \lim_{k\to \infty,k\in K} \zeta^k=0,\label{assert_D}\\
        && \lim_{k\to\infty,k\in K}\widetilde w^k_i =\bar w,\quad\mbox{for all $i=1,\dots, P$}\label{assert_A_0}\\
        && \lim_{k\to\infty,k\in K} d_k = -\nabla f(\bar w).\label{assert_B_0}\\
        && {\lim_{k\to\infty,k\in K} \tilde f^k =  f(\bar w).\label{assert_C_0}}
    \end{eqnarray}

\begin{proof}
Recalling that $\{w^k\}$ is limited and that $\lim_{k\rightarrow0}\zeta_k = 0$, \eqref{assert_C} and \eqref{assert_D} follow by definition.

Then, given a permutation $I^k = \{h_1,h_2,\dots,h_P\}$, consider the sequence $\| \widetilde w^k_{h_i}  - w^k \|$. 
Recalling that $\widetilde w^{k}_{0} = w^k$ follows by definition of \ref{alg:innercycle1} that:
$$ \widetilde w^k_{h_1} = w^k - \zeta^k \nabla f_{h_1} (w^k) 
$$
which implies that:
\begin{equation}
\label{eq:induction_hp}
    \| \widetilde w^k_{h_1} - w^k \| \le \zeta^k \| \nabla f_{h_1} (w^k)  \|
\end{equation}

For $i>1$ we can write:
\begin{equation}
    \| \widetilde{w}^k_{h_i} - w^k \| = \zeta^k \left\| \sum_{j=1}^i \nabla f_{h_j} (\widetilde{w}^k_{h_{j-1}}) \right\| = CO
\end{equation}
\begin{equation}
   CO \leq \sum_{j=1}^i \left\| \nabla f_{h_j} (\widetilde{w}^k_{h_{j-1}}) + \nabla f_{h_j} (w^k) - \nabla f_{h_j} (w^k) \right\|
   \label{eq:eq_di_supporto1}
\end{equation}

Exploiting the triangular inequality, \eqref{ass:lipschtizagradient}, and \eqref{ass:nablabounded}, we have that:
\begin{equation}
   \text{Eq. (\ref{eq:eq_di_supporto1})} \le \zeta^k \sum_{j=1}^i L \| \widetilde w^k_{h_{j-1}} - w^k \| + \| \nabla f_{h_j} (w^k )\| 
   \label{eq:eq_di_supporto2}
\end{equation}

This proves that:
\begin{equation}
\label{eq:IC_w_bounded}
\| \widetilde w^k_{h_i} - w^k \| \le \text{Eq. (\ref{eq:eq_di_supporto2}) for all $i=1,\dots,P$}
\end{equation}
Hence, we can now reason by induction. If $i=1$, then, taking the limit for $k\rightarrow \infty$ on \eqref{eq:induction_hp} and recalling that $\zeta^k \rightarrow 0$, we get:
$$ \lim_{k\rightarrow\infty, k \in K} \| \widetilde w^k_{h_1} - w^k \|  = 0
$$
Assuming that this is true for $i=1,2,\dots,j-1$, thanks to the boundedness of $\{w^k\}$, \eqref{assert_A_0} is proved.

Let's now consider the direction given by Algorithm \ref{alg:innercycle1}:
$$ d^k = - \sum_{i=1}^P \nabla f_{h_i} (\widetilde w^k_{h_{i-1}})
$$
Taking the limit for $k \rightarrow \infty$ and using \eqref{assert_A_0} and Assumption \ref{ass:lipschtizagradient}, \eqref{assert_B_0} is proved.
In fact, we have:

$$ \lim_{k\rightarrow\infty, k \in K} d^k = - \sum_{i=1}^P \nabla f_{h_i} (\Bar w ) = - \nabla f(\Bar w)
$$

Eventually, using the continuity of $f(w)$ and \eqref{assert_A_0}, we have that
$$ \lim_{k\to\infty,k\in K} \tilde f^k = \lim_{k\to\infty,k\in K} \sum_{i=1}^P f_{h_i} (\widetilde w^k_{h_{i-1}}) =  \sum_{i=1}^P f_{h_i} (\Bar w) = f( \Bar w)
$$
which concludes the proof of \eqref{assert_C_0}

\end{proof}
\end{proposition}

\subsection{Derivative-Free Line-search}

DFL (Algorithm \ref{alg:EDFL2_cma1}) performs an extrapolation procedure similar to the one presented in \citep{coppola2023cma,liuzzi2022convergence}.
The step-size is increased by dividing the current learning rate by the hyper-parameter $\delta \in (0,1)$ until the condition at Step 7 holds.
However, in order to minimize the number of evaluations of the entire objective $f(w)$, we tailored the line-search to support the use of any coercive model $\psi$ to approximate $f$.

In our implementation, given a permutation $I_k = \{h_1,\dots,h_P\}$, $\psi$ is computed using a subset of the available samples, such that:
\begin{equation}\label{eq:psi}
    \psi(w) = \sum_{i=1}^p f_{h_i}(w) \text{ with $p < < P$ }
\end{equation}
The sufficient decrease condition at Step 13 (Algorithm \ref{alg:EDFL2_cma1}) is still evaluated on the true objective to deterministically prove the global convergence.
Furthermore, at the beginning of DFL, the current learning rate $\zeta^k$ is scaled by a factor $\eta \in (0,1)$ to achieve more stability than in \citep{liuzzi2022convergence}, where the line-search happens to fail when $\zeta^k$ is still too large.

 \begin{algorithm}[h!]
\caption{\texttt{Derivative-Free Linesearch (DFL) }}
\label{alg:EDFL2_cma1}
\footnotesize
\begin{algorithmic}[1]
\STATE  Input $(\tilde f^k,w^k,d^k,\zeta^k;\gamma,\delta,\psi(\cdot))$: $w^k\in \R^n,d^k\in \R^n,\zeta^k>0$, $\gamma \in (0,1), \delta\in(0,1), \eta \in (0,1)$
\STATE {Set $j = 0, \alpha = \zeta^k \eta, f_j = \tilde f^k$}
\STATE Compute $f(w^k)$
\IF {${\tilde f^k}> {f(w^k)}-\gamma\alpha\|d^k\|^2$}
    \label{edfl:step4}    \STATE  Set $\alpha^k = 0$ and \textbf{return} $\alpha^k, f_j$
\ENDIF 
\WHILE {${\psi(w^k +(\alpha / \delta) d^k)}\leq \min\{f(w^k)-\gamma \alpha\|d^k\|^2, {f_j}\}$}
     \label{edfl:step7}   
     \STATE {Set $f_{j+1} = \psi (w^k +(\alpha / \delta) d^k)$}
     \STATE {Set $j = j + 1$}
     \STATE {Set $\alpha = \alpha / \delta$}
\ENDWHILE
\STATE Compute $f(w^k + \alpha d^k)$
\IF{$f(w^k + \alpha d^k) \le f(w^k) - \gamma\alpha\|d^k\|^2$}
\STATE  Set $\alpha^k = \alpha$ and \textbf{return} $\alpha^k, {f (w^k +\alpha^k d^k)}$
\ELSE
\STATE  Set $\alpha^k = 0$ and \textbf{return} $\alpha^k, f_0$
\ENDIF
\end{algorithmic}
\end{algorithm}

In the following proposition, we show an important property of the Algorithm \ref{alg:EDFL2_cma1}, which is also crucial to prove the convergence of F-CMA.

\begin{proposition}[DFL well-defined]
\label{prop:EDFL}
Suppose that $\psi(\cdot)$ is a coercive approximation of $f(w)$.
Then, Algorithm \ref{alg:EDFL2_cma1} is well-defined, and it returns within a finite number of steps a scalar $\alpha_k$ such that:
\begin{equation} \label{eq:edfl_property}
    f (w^k + \alpha^k d^k) \le f(w^k) - \gamma \alpha^k \|d^k\|^2
\end{equation}
\begin{proof}
First, we show by contradiction that the EDFL cannot cycle an infinitely number of times. 
If this was the case, we would have that, for all $j=1,2,\dots$:
$$ \psi (w^k + \frac{\zeta^k \eta}{\delta^j} d^k) \le f(w^k) - \gamma  \frac{\zeta^k \eta}{\delta^j}\|d^k\|^2
$$

Taking the limit for $j \rightarrow \infty$ and being $\frac{\zeta^k \eta}{\delta^j}\rightarrow \infty$, we would get a contradiction with the compactness of the level sets of $\psi$ and with the boundedness of $f(w)$.

To prove \eqref{eq:edfl_property}, we first consider that it trivially holds if the EDFL fails (Step 4) and $\alpha^k = 0$. 
Conversely, if the condition at Step 7 is verified at least once, we can select a scalar $\alpha > 0$, only if \eqref{eq:edfl_property} holds (Step 13).
\end{proof}
\end{proposition}

\subsection{Convergence of F-CMA} 

F-CMA (Algorithm \ref{alg:CMA1}) is based on the random reshuffling iteration.
The assumptions stated at the beginning of this section, jointly with the algorithm conditions at Steps 6, 9, and 13, ensure the global convergence to a stationary point for smooth objective functions.

At each epoch $k$, after performing Algorithm \ref{alg:innercycle1} and having a tentative point $\tilde w^k$, a tentative direction $d^k$, and an estimate of the objective function in the point $\tilde w^k$, $\tilde f^{k+1}$ given by \eqref{eq:f_tilde}, F-CMA checks at Step 6, whether there has been an improvement by at least a factor $\gamma \zeta^k$, being $\zeta^k$ the decreasing learning rate.
Until the condition at Step 6 is satisfied, which often occurs in the first epochs of training, F-CMA produces the same points that would have been produced by RR.
When the condition is no longer satisfied, a safeguard rule on the norm of the direction is checked.
If the norm is too small with respect to the hyper-parameter $\tau$ (condition at Step 9), the learning rate $\zeta^k$ is decreased by the factor $\theta$ but, if still in the level set of $f(w)$, the tentative point is accepted.
Otherwise, DFL (Algorithm \ref{alg:EDFL2_cma1}) is executed, using any suitable model $\psi(\cdot)$ of the true objective $f(w)$.
The line-search returns an Armijo-like step-size \citep{armijo1966minimization,grippo1988global,grippo2007} and, according to condition at Step 13, the learning rate can be updated as follows: decreased, set to its returned value, or driven down to zero for the current epoch in case we end up out of the level set (conditions at steps 14.3 or 16.2).

\begin{algorithm}[h!]
\footnotesize
\begin{algorithmic}[1]
    \STATE  Set $ \zeta^0 >0, \theta\in(0,1), \tau > 0$, $\gamma\in(0,1), \delta\in(0,1), \eta \in (0,1), \alpha_{min}>0$
    \STATE  Let $w^0\in \R^n$
    \STATE  Compute $f(w^0)$ and set $\tilde f^{0} =  f(w^0)$ and ${\phi^0 = \tilde f^0}$
    \FOR {$k=0,1\dots $}
        
        \STATE  Compute $(\widetilde w^k, d^k, {\tilde f^{k+1}})$ = \texttt{RR}($w^k,\zeta^k$)
   \IF {{ ${\tilde f^{k+1}} \leq \min \{{\phi^{k}} - \gamma\zeta^k, f(\omega^0)\} $}}  
    \STATE  {Set $\zeta^{k+1} = \zeta^k$} \label{cma2:step11} {and $\alpha^k = \zeta^k$} and ${\phi^{k+1} = \tilde f^{k+1}}$
    \ELSE 
    \IF {$\|d^k\| \leq \tau\zeta^k$}
        \STATE  \label{cma2:step9} Set $\zeta^{k+1}=\theta \zeta^k$
        and {$\alpha^k=\begin{cases}
         \zeta^k & \text{if }{\tilde f^{k+1}}\le f(w^0)\\
        0 & \text{otherwise}\\
        \end{cases}$} and $\phi^{k+1} = \phi^k$
        
    \ELSE 
         \STATE  $(\widetilde\alpha^k, {\hat f^{k+1}}) =\texttt{DFL}(w^k,d^k,\zeta^k;\gamma,\delta,\eta) $
         \IF  {$\widetilde\alpha^k\|d^k\|^2 \leq\tau\zeta^k$}
            \STATE  \label{step19_CMA2}Set $\zeta^{k+1}=\theta \zeta^k$  and $\alpha^k=\begin{cases} 
          \widetilde\alpha^k & \text{if } \widetilde\alpha^k > 0 \text{ {and} } {\hat f^{k+1} \le f(w^0)} \\
          \zeta^k & \text{if } \widetilde\alpha^k=0 \text{ and }{\tilde f^{k+1}}\le f(w^0)\\
          0 & \text{otherwise}\\
        \end{cases}$ 
                   
         \ELSE 
            \STATE  Set $\zeta^{k+1}=\max\{\widetilde\alpha^k, \alpha_{min}\},$ \label{step18_CMA2}
            and $\alpha^k = \begin{cases} 
          \widetilde\alpha^k & \text{if } \widetilde\alpha^k > 0 \text{ {and} } { \hat f^{k+1} \le f(w^0)} \\
          0 & \text{otherwise}\\
        \end{cases}$
         \ENDIF 
    \STATE  {Set $\phi^{k+1} = \min \{\hat f^{k+1}, \tilde f^{k+1}, \phi^k\}$}
    \ENDIF 
    \ENDIF 
    \STATE  Set $w^{k+1} = w^k + \alpha^k d^k$ 
   \ENDFOR 
\end{algorithmic}
\caption{{\texttt{ F-CMA }}}
\label{alg:CMA1}
\end{algorithm}

Proving the global convergence of F-CMA to a stationary point for $f(w)$ is straightforward once Proposition \ref{prop:limitRR} can be applied.
In fact, the proofs are mostly by contradiction, and they are all based on the analysis of the possible outcomes of each epoch $k$ of Algorithm \ref{alg:CMA1}, i.e.: 
\begin{itemize}
    \item[a)] tentative point accepted and learning rate unchanged (Step 7);
    \item[b)] direction too small (condition at Step 9) and learning rate decreased, then tentative point accepted (Step 10.1) or rejected (Step 10.2);
    \item[c)] perform DFL (Step 12), direction too small (Step 13), learning rate decreased, then DFL point accepted (Step 14.1) or tentative point accepted (Step 14.2) or both rejected (Step 14.3);
    \item[d)] perform DFL (Step 12), DFL point accepted (Step 16.1) if it is in the level set, otherwise rejected (Step 16.2).
\end{itemize}

The key steps to prove that Proposition \ref{prop:limitRR} can be applied are the following:
\begin{itemize}
    \item[i)] Prove, in Lemma \ref{lemma:bound_f_tilde}, that the difference between the estimate $\tilde f$ and the real objective function $f(w)$ is bounded by a certain fixed threshold.
    \item[ii)]  Prove that the sequence of points $\{w^k\}$ produced by F-CMA is limited (Proposition \ref{prop:limitedw}).
    \item[iii)] Prove that the learning rate goes to zero, i.e., $\zeta^k \rightarrow 0$ (Proposition \ref{zetazero}).
\end{itemize}

More in detail, we firstly need to prove that the difference between the approximation of the objective function $\tilde f^{k+1}$ and the real value $f(w^k)$ at the beginning of the epoch is bounded by a certain fixed threshold.
In the following lemma, we show that this threshold exists and depends on the constants $C$ and $M$, on the number of samples (or batch of samples) $P$, as well as on the Lipschitz constant of $f(w)$.

\begin{lemma}[$| \tilde f^{k+1} - f(w^k) |$ bounded]
\label{lemma:bound_f_tilde}
The following relation between the estimate of the objective function $\tilde f^k$ and the real objective function $f(w^k)$ holds for every iteration $k=0,1,\dots$ of Algorithm \eqref{alg:CMA1}.

\begin{equation} \label{eq:bound_f_tilde}
    | \tilde f^{k+1} - f(w^k) | \le P^2 L_{f} \zeta^k (CM + D)
\end{equation}

where $L_f$ is the Lipschitz-constant of $f(w)$.
\begin{proof}
    To avoid an abuse of notation, the proof is written in the case of the trivial random permutation $I^k = \{1,2,\dots,P\}$.

    Let's consider the difference between the estimate and the real value of the $p$-th term of $f(w)$. Recalling Assumption \ref{ass:lipschtizagradient} and naming $L_f = \max_{p=1,\dots,P} L_{f_p}$ we can write:
    $$ |f_p(\tilde w^k_{p-1}) - f_p(w^k) | \le L_f \| \tilde w^k_{p-1} - w^k\| =$$ 
    $$L_{f} \zeta^k \| \sum_{j=1}^p \nabla f_j (\tilde w^k_{j-1}) \| \le 
     L_{f} \zeta^k p C (\| \nabla f (\tilde w^k_{j-1}) \| + D) \le $$
     
     $$\le L_{f} \zeta^k p (C M + D)
    $$

    where for the last inequalities we exploited Assumption \ref{ass:nablabounded} and the hypothesis on $\| \nabla f(w) \|$.
    Now we can rewrite:
    $$ | \tilde f^{k+1} - f(w^k) |  = | \sum_{p=1}^P f_p(\tilde w^k_{p-1}) - f_p(w^k)| \le
    $$
    
    $$  \le \sum_{p=1}^P | f_p(\tilde w^k_{p-1}) - f_p(w^k) | \le \sum_{p=1}^P L_{f} \zeta^k p (C M + D) =  $$
    $$= L_{f} \zeta^k (C M + D) \frac{P(P-1)}{2} \le L_{f} \zeta^k (C M + D) P^2
    $$
    which proves the thesis.
\end{proof}
\end{lemma}

We have now collected all the elements to prove, in the following proposition, that the sequence of points produced by F-CMA is limited.

\begin{proposition}[$\{w^k\}$ limited]
\label{prop:limitedw}
Let $\{w^k\}$ be the sequence of points produced by the Algorithm \ref{alg:CMA1}.
Then, $\{w^k\}$ is limited.

\begin{proof}
The proof begins showing that for each $k=0,1\dots$ at least one of the following conditions holds:
\begin{itemize}
    \item[i)] $\tilde f^{k+1} \le f(w^0)$
    \item[ii)] $f(w^k) \le f(w^0)$
    \item[iii)] $w^{k+1} = w^k$
\end{itemize}

By definition of Algorithm \ref{alg:CMA1},  for every iteration $k$, we have that one of the following cases happens:
\begin{enumerate}
    \item[a)] Step 7 is executed, i.e., we move to a new point such that $\tilde f^k \le \min \{\phi^{k-1} - \gamma \zeta^k, f(\omega^0) \}$.
    \item[b)] Step 10 is executed. We move to a point such that $\tilde f^k \le f(\omega^0)$, or we set $\alpha^k = 0$, i.e., we do not move from $\omega^k$.
    \item[c)] Step 14 is executed. After the line-search, we move to a point such that either $\tilde f^k \le f(\omega^0)$ or  $f(\omega^k + \tilde \alpha^k d^k) = f(\omega^{k+1}) \le f(\omega^0)$ or we do not move if neither of the conditions is satisfied.
    \item[d)] Step 16 is executed. After the line-search we move to a point such that $f(\omega^k + \tilde \alpha^k d^k) = f(\omega^{k+1}) \le f(\omega^0)$ or we do not move. 
\end{enumerate}

In case a) at least i) holds. 
In case b) either i) or iii) holds. 
In case c) either i) or ii) holds if the line-search learning rate is satisfied otherwise iii) holds.
In case d), again, ii) holds if the line-search learning rate is satisfied, otherwise iii) holds.

Now, suppose by contradiction that the sequence of points $\{w^k\}$  produced by the Algorithm \ref{alg:CMA1} diverges, i.e., an infinite set of index $k \subseteq \{0,1,2,\dots\}$ exists such that:

\begin{equation}\label{normtoinfty}
\displaystyle \lim_{\substack{k \rightarrow \infty \\ k \in K}} \vert \vert \omega^k \vert \vert = \infty
\end{equation}

Consider the following sets of indexes:
\begin{itemize}
    \item $K_1 = \{k = 0,1,\dots: \tilde f^{k+1} \le f(w^0) \}$
    \item $K_2 = \{k = 0,1,\dots: \tilde f(w^k) \le f(w^0) \}$
    \item $K_3 = \{k = 0,1,\dots: w^{k+1} = w^k \}$
\end{itemize}

We first show by contradiction that $K \cap (K_1 \cup K_2)$ must be finite. 

For any $k$ such that i) holds, recalling \eqref{eq:bound_f_tilde}, we have that:
$$ f(w^k) \le \tilde f^{k+1} + P^2 L_f \zeta^k (C M + D) \le f(w^0) +  P^2 L_f \zeta^k (C M + D)
$$
If $K \cap (K_1 \cup K_2)$ is not finite, then taking the limit for $k\rightarrow\infty$ with $k \in K \cap (K_1 \cup K_2)$ on both sizes and recalling that $f(w)$ is coercive (assumption \eqref{ass:compact}) and that $\{\zeta^k\}$ is by definition a non-increasing sequence, we get the following contradiction:
$$ \infty \le f(w^0) +  P^2 L_f \bar \zeta (C M + D) < \infty
$$
Conversely, for any $k$ such that ii) holds, if $K \cap (K_1 \cup K_2)$ is not finite, we immediately have a contradiction with assumption \eqref{ass:compact}, since we would get that $\|w^k\| \rightarrow \infty$ but $f(w^k) \le f(w^0)$, meaning that the level set is not compact.

Hence, since $K \cap (K_1 \cup K_2)$ must be finite, there must be an infinite set of index $\tilde K \subseteq K_3 \setminus (K_1 \cup K_2) $ such that:
\begin{equation}\label{eq:norminftyktilde}
    \displaystyle \lim_{\substack{k \rightarrow \infty \\ k \in (\tilde K \cap K)}} \vert \vert \omega^k \vert \vert = \infty
\end{equation}

But here we encounter again a contradiction.
In fact, let's consider an index $\bar k \in  (\tilde K \cap K)$. 
If for all $m=\{1,2,\dots\}$ 
$\bar k + m \in  (\tilde K \cap K)$, then  we get a trivial contradiction with $\| w^k\| \rightarrow \infty$.
Hence, to ensure \eqref{normtoinfty}, there must be an index $\tilde k > \bar k$ such that $w^{\tilde k} \neq w^{\bar k}$, which means such that iii) does not hold.
However, we have proved that if iii) does not hold, at least one between i) and ii) must hold, which means ${\tilde k} \in (K_1 \cup K_2)$. But this is, by definition of $\tilde K$, in contradiction with $k \in  (\tilde K \cap K)$.
\end{proof}
\end{proposition}

Finally, the following proposition proves that the learning rate goes to zero.
\begin{proposition}[$\zeta^k \rightarrow 0$]
\label{zetazero}
Let $\{\zeta^k\}$ be the sequence of steps produced by Algorithm \ref{alg:CMA1}, then
\[
 \lim_{k\to\infty}\zeta^k = 0.
\]
\end{proposition}
\begin{proof}
In every iteration, either $\zeta^{k+1} = \zeta^k$ or $\zeta^{k+1} = \theta\zeta^k < \zeta^k$. Therefore, the sequence $\{\zeta^k\}$ is monotonically non-increasing. Hence, it results
\[
\lim_{k\to\infty} \zeta^k = \bar\zeta \geq 0.
\]
Let us suppose, by contradiction, that $\bar\zeta > 0$. 
If this were the case, there should be an iteration index $\bar k\ge 0$ such that, for all $k\ge \bar k$, $\zeta^{k+1} = \zeta^k = \bar\zeta \ge \alpha_{min}$. 
Namely, for all iterations $k\ge \bar k$, steps 7 or 16 are always executed.

Let us now denote by
\begin{eqnarray*}
  K^\prime & = & \{k:\ k\geq\bar k,\ \mbox{and step 7 is executed}\}\\
  K^{\prime\prime} & = & \{k:\ k\geq\bar k,\ \mbox{and step 17 is executed}\}  
\end{eqnarray*}

Let us first prove that  $K^\prime$ cannot be infinite. If this was the case, we would have that infinitely many times
\[
   \gamma\bar\zeta \leq \phi^k - \tilde f^{k+1} {= \phi^k - \phi^{k+1}}
\]
However, by the instructions of the algorithm $\{\phi^k\}$ is a monotonically non increasing sequence. 
Also, by Assumption \ref{ass:fiproperties}, for all $k$, $\phi^k\geq 0$. 
This means that: 
\[
\lim_{k\to\infty,k\in K'}\phi_k = \lim_{k\to\infty,k\in K'}\phi^{k+1} = \bar\phi.
\]
That is to say that 
\[
\lim_{k\to\infty,k\in K'}\phi^k-\phi^{k+1} = 0
\]
which is in contrast with $\bar\zeta > 0$.
\par\smallskip

Thus, an index $\hat k$ exists such that, $k\in K''$ for $k\geq\hat k$. 
Thanks to Proposition \ref{prop:EDFL}, for $k\geq\hat k$, we have
\begin{equation}\label{seqfw}
 f(w^{k+1}) \leq f(w^k+\alpha^kd^k) \leq {f(w^k)} -\gamma\alpha^k\|d^k\|^2,
\end{equation}
that is the sequence  $\{f(w^k)\}$ is definitely monotonically non-increasing and bounded below since $f(w) \geq 0$. Hence 
\[
\lim_{k\to\infty}f(w^k) = \lim_{k\to\infty} f(w^{k+1}) = \bar f.
\]
Then, 
taking the limit, we obtain:
\[
\lim_{k\to\infty} \alpha^k\|d^k\|^2 = 0.
\]
But then, for $k\geq \hat k$ sufficiently large, it would happen that 
\[
 \alpha^k\|d^k\|^2 \leq \tau\bar\zeta
\]
which means that step 10 would be executed, setting $\zeta^{k+1} = \theta\bar\zeta$ thus decreasing $\zeta^{k+1}$ below $\bar\zeta$, which contradicts our initial assumption and concludes the proof. 
\end{proof}

Following the last proof, it is now possible to show that F-CMA converges to a stationary point, i.e., that $\lim_{k\rightarrow\infty} \inf_k \| \nabla f(w^k) \| = 0$.

\begin{proposition}[Global convergence of F-CMA]
\label{prop:convergence}
    Let $\{w^k\}$ be the sequence of points produced by the Algorithm \ref{alg:CMA1}. Then, $\{w^k\}$ admits limit points, and at least one of them is a stationary point for $f(w)$.
\begin{proof}
    $\{w^k\}$ is limited by Proposition \ref{prop:limitedw}, thus it admits limit points.
    Furthermore, recalling Proposition \ref{zetazero}, the set of index:
    $$ K = \{k=0,1,\dots: \zeta^{k+1} = \theta \zeta^k \}
    $$
    must be infinite and, by Proposition \ref{prop:limitRR}, there exists an infinite set $\bar K \subseteq K$ such that 
    $$\displaystyle\lim_{k\rightarrow\infty, k \in \bar K} w^k = \bar w$$
    and
    $$\displaystyle\lim_{k\rightarrow\infty, k \in \bar K} d^k = - \nabla f(\bar w)$$

    Now, recalling conditions at Step 9 and at Step 13, we have that for all $k \in K$ either $\|d^k\|^2 \le \tau \zeta^k$, or $\tilde \alpha^k \| d^k\| \le \tau \zeta^k$.
    Hence, we can partition $\bar K$ as follows:
    \begin{eqnarray*}
  \bar K^\prime & = & \{k:\ \mbox{condition at step 9 is satisfied}\}\\
  \bar K^{\prime\prime} & = & \{k:\ \mbox{condition at step 13 is satisfied}\}  
\end{eqnarray*}
At least one between $\bar K^\prime$ and $\bar K^{\prime\prime}$ must be infinite.

If $K^\prime$ is infinite, then:
$$ \displaystyle\lim_{k\rightarrow\infty, k \in \bar K^\prime} d^k = - \nabla f(\bar w) \le \displaystyle\lim_{k\rightarrow\infty, k \in \bar K^\prime} \tau\zeta^k = 0
$$
and the thesis is proved.

If $\bar K^{\prime\prime}$ is infinite, then:
$$ \displaystyle\lim_{k\rightarrow\infty, k \in \bar K^\prime} d^k \tilde\alpha^k \le \displaystyle\lim_{k\rightarrow\infty, k \in \bar K^\prime} \tau\zeta^k = 0
$$
Recalling that $\tilde \alpha^k \le 0$ and using Assumption \ref{ass:nablabounded}, this means:
$$ \displaystyle\lim_{k\rightarrow\infty, k \in \bar K^\prime} d^k \tilde\alpha^k \le \displaystyle\lim_{k\rightarrow\infty, k \in \bar K^\prime} PC\nabla f(w^k) \tilde\alpha^k = 0
$$
If $\nabla f(w^k) \rightarrow 0$, the thesis is proved. 
Otherwise, it must hold that $ \tilde\alpha^k\rightarrow 0$.

Recalling Proposition \ref{prop:EDFL}, we have that, for all $k \in \bar  K^{\prime\prime}$,
$$ f(w^k + \tilde\alpha^k d^k) \le f(w^k) - \gamma\tilde\alpha^k\|d^k\|^2
$$

Applying the Mean Value theorem for all $k \in \bar  K^{\prime\prime}$, we know that there exists a scalar $\xi^k \in (0,\tilde\alpha^k)$ such that:
$$ \frac{f(w^k + \xi^k d^k) -  f(w^k) }{\xi} = \nabla f(w^k)^T d^k \le \gamma \|d^k\|^2 
$$
Taking the limit for $k \rightarrow\infty, k\in K^{\prime\prime}$, we obtain:
$$   \| \nabla f(\bar w) \|^2 \le \gamma \| \nabla f(\bar w) \|^2
$$
Recalling that $\gamma \in (0,1)$, this is possible only if $\| \nabla f(\bar w) \| = 0$, which proves the proposition.
\end{proof}
\end{proposition}

\subsection{Implementation strategy}

As introduced in Section \ref{sec:intro}, two of our key contributions concern implementation solutions to improve the proposed method efficiency.
The first one, embedded in the optimizer, is the possibility to enable the gradient clipping forcing the norm of the gradient to stay bounded.
Although this process is just an in-place alteration of the gradient, it has other efficiency benefits when applied to non-adaptive methods \citep{chen2020understanding,qian2021understanding}.
The second one is an early-stopping rule depending on the value of the current learning rate $\zeta^k$.
In F-CMA, one can set the optional hyper-parameter $\varepsilon$ (reported in table \ref{tab:hpFCMA}) to a certain threshold.
When $\zeta^k$ falls below this threshold, the optimizer returns a stop signal, and the training is interrupted.
Our computational experience shows that this procedure results in a great saving of computational effort while achieving competitive accuracy results.

\subsection{Hyper-parameters} 

As any other optimizer, F-CMA has hyper-parameters that can be tuned during a pre-training grid-search phase.
For the sake of simplicity, we provide in Table \ref{tab:hpFCMA} the default value of each hyper-parameter and a summary of its role.

\begin{table}[h!]
    \centering
    \footnotesize
    \caption{Hyper-parameters (HPs) default values and description for F-CMA}
    \begin{tabularx}{\linewidth}{c|c|l}
        \hline
        HPs & Value & Description \\
        \hline
        $\zeta^0$ & 0.05 & Initial learning rate \\
        $\theta$ & 0.75 & Learning rate decrease factor \\ 
        $\tau$ & 0.01 & Controls the norm of the direction \\
        $\gamma$ & 0.01 & Controls the sufficient decrease of $\tilde f$ \\
        $\delta$ & 0.90 & Step-size increase factor within DFL \\
        $\eta$ & 0.50 & Scaling factor of the learning rate before DFL \\
        $\alpha_{min}$ & $10^{-10}$ &  Minimal value allowed for the $\zeta^k$ if  Step 13 does not hold \\
        $\varepsilon$ & $10^{-10}$ & If $\zeta^k$ is lower, the training can be stopped \\
        \hline
    \end{tabularx}
    \label{tab:hpFCMA}
\end{table}

\section{Experimental setup}
\label{sec:expsetup}
This section gives a detailed description of the experimental setup, including training hyper-parameters, benchmark datasets, and evaluation metrics.

F-CMA has been implemented using PyTorch deep learning API.
In order to investigate the convergence behavior and estimation performances achieved by F-CMA, we select two of the most commonly used reference benchmark datasets for image classification, i.e., CIFAR10 and CIFAR100 \citep{krizhevsky2009learning}, and compare over eight optimizers and six well-known neural networks.
Each dataset is composed of $60K$ images, using the common train/val split composed of $50K$/$10K$ images, at a resolution of $32 \times 32$ pixels with $10$ and $100$ classes, respectively.
More in detail, we select as reference optimizers the original CMAL configuration, Adam, Addamax, AdamW, SGD, Adagrad, Nadam, and Radam using their default hyperparameters, i.e., the ones reported in their original papers with the exception of the learning rate of SGD that has been reduced to $1e^{-2}$ due to the lack of convergence in its original settings.
Moreover, we compare the optimizers performances over five convolutional and a single vision transformer neural networks, precisely, we chose as reference models ResNet-18, ResNet-50, ResNet-152 \citep{he2016deep}, Wide ResNet-50 \citep{zagoruyko2016wide}, MobileNet V2 \citep{sandler2018mobilenetv2} and Swin-B \citep{liu2021swin}.
All the models have been initialized on the ImageNet \citep{deng2009imagenet} dataset and sequentially trained with each optimizer, i.e., in 54 different combinations, with a batch size of $128$ images for $250$ epochs ($N_{ep}$) over a single NVIDIA RTX 3090 GPU.

Once the training phase is concluded, we evaluate each model-optimizer combination based on accuracy (Acc) values to determine the architecture's performance on the classification tasks.
Moreover, we also report in Table \ref{tab:results} the number of epochs ($N_{ep}$) needed to perform a complete train, the average time needed to perform a single epoch (T/Ep) measured in seconds, and the epoch in which the highest value of the validation accuracy has been reached ($k^*$).

\begin{table*}[htbp]
    \centering
    \caption{Model-optimizer performances over classification tasks. The proposed method is highlighted in gray, the best results are in bold, and the second bests are underlined.}
    \label{tab:results}
    \footnotesize
    \begin{tabular}{ll | cccc | cccc }
        \hline
        \multirow{2}*{\textbf{Net}} & \multirow{2}*{\textbf{Optimizer}} & \multicolumn{4}{c|}{\textbf{CIFAR 10}} &\multicolumn{4}{c}{\textbf{CIFAR 100}} \\
         & & Acc $\uparrow$ & T/Ep $\downarrow$ & $N_{ep}$ & $k^*$$\downarrow$ &  Acc $\uparrow$ & T/Ep $\downarrow$& $N_{ep}$ & $k^*$$\downarrow$ \\
        \hline
                \multirow{9}{*}{\rotatebox{90}{\textbf{ResNet-18}}} 
            & Adam       & 81.67 & \underline{6.46} & 250 & 139 
                         & 52.30 & 6.55  & 250 & 31\\
            & Adamax    & 82.69 & 7.75 & 250 & 228 
                        & 53.14 & 7.79 & 250 & 86\\
            & AdamW     & 81.24 & 6.52 & 250 & \textbf{78}  
                         & 52.46 & 6.49 & 250 & 35\\
            & SGD        & 79.27 & \textbf{5.96} & 250 & 201  
                         & 51.82 & \textbf{5.85} & 250 & 139\\
            & Adagrad    & 82.02 & 6.51 & 250 & \underline{106}  
                         & 53.03 & \underline{6.38} & 250 & 250\\
            & Nadam      & 81.27 & 6.63 & 250 & 181  
                         & 51.33 & 6.65 & 250 & \textbf{20} \\
            & Radam      & 82.02 & 6.62 & 250 & 114  
                         & 52.78 & 6.69 & 250 & \underline{29} \\
            & CMAL       & \underline{82.83} & 7.45 & 250 & 119 
                          & \underline{54.95} & 7.64 & 250 & 179\\
            & \cellcolor{lightgray}\textbf{F-CMA}     & \cellcolor{lightgray}\textbf{83.43} & \cellcolor{lightgray}6.52 & \cellcolor{lightgray}109 & \cellcolor{lightgray}\textbf{78} 
                          & \cellcolor{lightgray}\textbf{56.64} & \cellcolor{lightgray}6.46 & \cellcolor{lightgray}119 & \cellcolor{lightgray}79 \\
        \hline
                \multirow{9}{*}{\rotatebox{90}{\textbf{ResNet-50}}} 
            & Adam       & 82.34 & 12.65 & 250 & 242 
                         & 49.76 & 12.67 & 250 & 102  \\
            & Adamax    & 83.64 & 15.45 & 250 & 142  
                         & 56.65 & 15.53 & 250 & \textbf{15} \\
            & AdamW     & 81.67 & 12.75 & 250 & \underline{77}  
                         & 52.02 & 12.76 & 250 & \underline{27} \\
            & SGD        & 81.74 & \textbf{11.95} & 250 & 227  
                         & 56.63 & \textbf{11.91} & 250 & 191 \\
            & Adagrad    & 83.01 & \underline{12.57} & 250 & 223  
                         & 50.91 & \underline{12.53} & 250 & 201 \\
            & Nadam      & 81.43 & 13.04 & 250 & 198  
                         & 49.17 & 13.05 & 250 & 81 \\
            & Radam      & 82.86 & 12.92 & 250 & 187  
                         & 54.76 & 12.91 & 250 & 37 \\
            & CMAL       & \underline{84.96} & 13.90 & 250 & 195 
                         & \underline{59.58} & 13.52 & 250 & 70 \\
            & \cellcolor{lightgray}\textbf{F-CMA}     & \cellcolor{lightgray}\textbf{85.55} & \cellcolor{lightgray}12.70 & \cellcolor{lightgray}93 & \cellcolor{lightgray}\textbf{75} 
                         & \cellcolor{lightgray}\textbf{60.51} & \cellcolor{lightgray}12.57 & \cellcolor{lightgray}96 & \cellcolor{lightgray}62 \\
        \hline
                \multirow{9}{*}{\rotatebox{90}{\textbf{ResNet-152}}} 
            & Adam       & 77.89 & 25.58 & 250 & 205  
                         & 46.78 & 25.79 & 250 & 226 \\
            & Adamax    & \underline{82.91} & 33.05 & 250 & \underline{139}  
                         & 54.51 & 33.11 & 250 & 47 \\
            & AdamW     & 79.57 & 25.91 & 250 & 239  
                         & 45.61 & 25.90 & 250 & 227 \\
            & SGD        & 80.94 & \textbf{23.77} & 250 & 243  
                         & \underline{55.53} & \textbf{23.77} & 250 & 205 \\
            & Adagrad    & 82.55 & 25.31 & 250 & 192  
                         & 45.08 & 25.24& 250 & \underline{41} \\
            & Nadam      & 80.83 & 26.53 & 250 & 202  
                         & 46.73 & 26.46 & 250 & 168 \\
            & Radam      & 81.72 & 26.34 & 250 & 245  
                         & 53.82 & 26.33 & 250 & 69 \\
            & CMAL       & 81.38 & 27.38 & 250 & \textbf{99}  
                         & 47.06 & 29.34 & 250 & \textbf{6} \\
            & \cellcolor{lightgray}\textbf{F-CMA}     & \cellcolor{lightgray}\textbf{83.98} & \cellcolor{lightgray}\underline{25.07} & \cellcolor{lightgray}105 & \cellcolor{lightgray}\textbf{99} 
                         & \cellcolor{lightgray}\textbf{58.90} & \cellcolor{lightgray}\underline{25.01} & \cellcolor{lightgray}111 & \cellcolor{lightgray}103  \\
        \hline
                \multirow{9}{*}{\rotatebox{90}{\textbf{Wide ResNet-50}}} 
            & Adam       & 82.20 & 13.77 & 250 & 186  
                         & 48.27 & 13.74 & 250 & 241 \\
            & Adamax    & 84.58 & 16.18 & 250 & 178  
                         & 56.79 & 16.31 & 250 & \textbf{53} \\
            & AdamW     & 81.78 & 13.91 & 250 & 114  
                         & 49.56 & 13.87 & 250 & 247 \\
            & SGD        & 83.06 & \textbf{12.30}  & 250 & 232  
                         & 57.96 & \textbf{12.35} & 250 & 221 \\
            & Adagrad    & 82.33 & 13.36 & 250 & 240  
                         & 49.44 & 13.31 & 250 & 173 \\
            & Nadam      & 81.10 & 14.16 & 250 & 231  
                         & 47.10 & 14.03 & 250 & 80 \\
            & Radam      & 83.37 & 14.25 & 250 & 137  
                         & 55.46 & 14.16 & 250 & 147 \\
            & CMAL       & \underline{85.18} & 14.57 & 250 & \textbf{61}  
                         & \underline{60.18} & 14.01 & 250 & 227 \\
            & \cellcolor{lightgray}\textbf{F-CMA}     & \cellcolor{lightgray}\textbf{85.93} & \cellcolor{lightgray}\underline{13.10} & \cellcolor{lightgray}85 & \cellcolor{lightgray}\underline{70}  
                         & \cellcolor{lightgray}\textbf{61.71} & \cellcolor{lightgray}\underline{12.99} & \cellcolor{lightgray}98 & \cellcolor{lightgray}\underline{59} \\
        \hline
                \multirow{9}{*}{\rotatebox{90}{\textbf{Mobilenet V2}}} 
            & Adam       & 81.33 & \underline{11.43} & 250 & 222  
                         & 52.98 & 11.46 & 250 & 141 \\
            & Adamax    & \textbf{83.50} & 14.47 & 250 & 84  
                         & 56.21 & 14.57 & 250 & \textbf{43} \\
            & AdamW     & 82.27 & 11.67 & 250 & 204  
                         & 53.16 & 11.55 & 250 & 90 \\
            & SGD        & 80.88 & \textbf{10.87} & 250 & 245  
                         & 54.99 & \textbf{10.79} & 250 & 248 \\
            & Adagrad    & 80.98 & 11.58 & 250 & \underline{89}  
                         & 52.41 & \underline{11.37} & 250 & \textbf{43} \\
            & Nadam      & 81.61 & 11.85 & 250 & 239 
                         & 47.48 & 11.85 & 250 & 193 \\
            & Radam      & 82.22 & 11.59 & 250 & 188  
                         & 55.25 & 11.68 & 250 & \underline{50} \\
            & CMAL       & 82.13 & 12.45 & 250 & 227  
                         & 54.90 & 12.55 & 250 & 229 \\
            & \cellcolor{lightgray}\textbf{F-CMA}     & \cellcolor{lightgray}\underline{83.01} & \cellcolor{lightgray}11.68 & \cellcolor{lightgray}78 & \cellcolor{lightgray}\textbf{41}  
                         & \cellcolor{lightgray}\textbf{57.64} & \cellcolor{lightgray}11.39 & \cellcolor{lightgray}122 & \cellcolor{lightgray}98 \\
            \hline
                \multirow{9}{*}{\rotatebox{90}{\textbf{Swin-B}}} 
            & Adam       & 85.68 & 55.96 & 250 & 249  
                         & 59.23 & 55.82 & 250 & 102 \\
            & Adamax    & 86.59 & 57.92 & 250 & 248  
                         & 62.47 & 57.74 & 250 & 241  \\
            & AdamW     & 81.53 & 56.42 & 250 & 238  
                         & 58.95 & 56.03 & 250 & 84 \\
            & SGD        & 87.07 & \textbf{53.29} & 250 & 239  
                         & 65.91 & \textbf{53.41} & 250 & 250 \\
            & Adagrad    & 86.65 & \underline{55.27} & 250 & 243 
                         & 63.93 & \underline{55.09} & 250 & 243 \\
            & Nadam      & 14.38 & 56.44 & 250 & -  
                         & 56.15 & 56.61 & 250 & 71 \\
            & Radam      & 85.30 & 56.63 & 250 & 178  
                         & 60.07 & 56.76 & 250 & \textbf{64} \\
            & CMAL       & \textbf{89.77} & 59.31 & 250 & \underline{94}  
                         & \underline{69.91} & 59.05 & 250 & 148 \\
            & \cellcolor{lightgray}\textbf{F-CMA}     & \cellcolor{lightgray}\underline{89.67} & \cellcolor{lightgray}55.32 & \cellcolor{lightgray}102 & \cellcolor{lightgray}\textbf{64} 
                         & \cellcolor{lightgray}\textbf{70.81} & \cellcolor{lightgray}55.16 & \cellcolor{lightgray}120 & \cellcolor{lightgray}\underline{70} \\
        \hline
    \end{tabular}
\end{table*}

\section{Computational experiments}
\label{sec:results}
In this section, we report the quantitative results achieved by our F-CMA when trained in combination with the six chosen architectures over the CIFAR10 and CIFAR100 classification datasets.  
As previously introduced, we compare F-CMA performances with eight other optimizers; the obtained results are reported in Table \ref{tab:results}.

Based on the reported results, it can be observed that in most cases, F-CMA achieves the highest estimation accuracy, with a maximum boost of up to $+5\%$ compared to other optimizers and a total training time of up to $-68\%$ shorter. 
More in detail, we note that in terms of the time needed to perform a single epoch, SGD is the fastest optimizer, due to its limited amount of operations. 
However, compared to F-CMA, SGD achieves an average boost of only $1$ second per epoch, and its convergence is usually slower, requiring an average of $+145$ additional epochs to reach $k^*$.
Furthermore, by observing the Acc and $k^*$ columns in the Table \ref{tab:results}, it can be noticed that F-CMA reaches higher accuracy values faster than all other optimizers in $\approx85\%$ of cases; consequently, the addition of the learning rate-driven early stopping rule significantly reduces the total training time.
More in detail, taking the Swin-B architecture as an example, it can be noticed that F-CMA achieves accuracy values comparable to CMAL, i.e., $-0.1\%$ and $+0.9\%$ respectively on CIFAR10 and CIFAR100 datasets with faster training time, i.e., a T/Ep equal to $-4.0$ and $-3.9$ seconds and a $k^*$ achieved in $-30$ and $-78$ epochs, leading to a reduction of $-36.5\%$ and $-55.8\%$ in the training time needed to converge ($k^*$).
Moreover, in this example, it can also be noticed that Nadam, with its default hyperparameters, is not able to converge ($k^* =$ -). 
Furthermore, by observing the cases in which the optimizers reach a $k^*$ value in a range lower than $40/50$ epochs, this phenomenon usually leads to low estimation accuracy values; for example, this behavior can be noticed in the following optimizer/architecture combinations: CMAL with ResNet-152 and Nadam with ResNet-18 on the CIFAR100 dataset.

Hence, we can assess that in the analyzed classification tasks, F-CMA exhibits a faster convergence behavior than well-established algorithms while attaining higher accuracy values when compared to the same deep learning model. 
These performances will also represent a promising advancement towards enhancing the efficiency of the training process, reducing its time and consequently lowering the energy consumption and the carbon footprint while avoiding architectural modification and/or potential performance losses \citep{ijcai2023p764, papa2024survey}.

\section{Conclusions and future works}
In this work, we introduce F-CMA, a new optimization algorithm capable of outperforming well-known optimizers in terms of convergence and computational efficiency.
Our quantitative findings, based on extensive analysis across many neural network architectures and benchmark datasets, validate F-CMA's theoretical convergence properties.  
Future research will investigate F-CMA capabilities in different types of architectures (e.g., diffusion models), as well as its behavior across other domains and tasks.

\section{Data availability and reproducibility}
The results presented in this paper are fully reproducible and the code is available in the public repository at \url{https://github.com/corradocoppola97/Fast_CMA}.

\section{Acknowledgements}
This study has been supported by ALCOR Lab, which provided the hardware resources to carry out the intensive computational tests, and by the following research projects:
\begin{itemize}
    \item Interpretable machine learning tools for health applications
    Progetti di Ricerca (Piccoli, Medi) - Progetti Medi
    Responsabile: PALAGI Laura - 1 Altro - 2 Doc./Ric. - 7 Dott./Ass./Spec.
    ID: RM1221816BAE8A79
    \item Optimization for Enhancing Machine Learning Models
    Progetti di Ricerca (Piccoli, Medi) - Progetti Medi
    Responsabile: PICCIALLI Veronica - 2 Altro - 2 Doc./Ric. - 4 Dott./Ass./Spec.
    ID: RM123188F479D981

\end{itemize}

\bibliographystyle{elsarticle-num} 
\bibliography{biblio.bib}

\end{document}